\documentclass[runningheads]{llncs}
\usepackage{graphicx}
\usepackage{amsmath}
\usepackage{amsfonts}
\usepackage{adjustbox}
\usepackage{xcolor}
\usepackage{pifont}
\usepackage{multirow}
\usepackage{booktabs}
\usepackage[T1]{fontenc}
\usepackage[hidelinks,colorlinks=true,linkcolor=blue,citecolor=blue]{hyperref}
\usepackage[capitalize]{cleveref}

\begin{document}

\title{CoMoTo: Unpaired Cross-Modal Lesion Distillation Improves Breast Lesion Detection in Tomosynthesis}
\titlerunning{CoMoTo}
\author{
Muhammad Alberb\inst{1} \and
Marawan Elbatel\inst{2} \and 
Aya Elgebaly\inst{1} \and 
Ricardo Montoya-del-Angel \inst{1} \and 
Xiaomeng Li\inst{2} \and 
Robert Martí\inst{1}
}
\authorrunning{Alberb et al.} 

\institute{
Computer Vision and Robotics Institute, University of Girona \and
The Hong Kong University of Science and Technology 
}

\maketitle

\begin{abstract}
Digital Breast Tomosynthesis (DBT) is an advanced breast imaging modality that offers superior lesion detection accuracy compared to conventional mammography, albeit at the trade-off of longer reading time. Accelerating lesion detection from DBT using deep learning is hindered by limited data availability and huge annotation costs. A possible solution to this issue could be to leverage the information provided by a more widely available modality, such as mammography, to enhance DBT lesion detection. In this paper, we present a novel framework, CoMoTo, for improving lesion detection in DBT. Our framework leverages unpaired mammography data to enhance the training of a DBT model, improving practicality by eliminating the need for mammography during inference. Specifically, we propose two novel components, Lesion-specific Knowledge Distillation (LsKD) and Intra-modal Point Alignment (ImPA). LsKD selectively distills lesion features from a mammography teacher model to a DBT student model, disregarding background features. ImPA further enriches LsKD by ensuring the alignment of lesion features within the teacher before distilling knowledge to the student. Our comprehensive evaluation shows that CoMoTo is superior to traditional pretraining and image-level KD, improving performance by 7\% Mean Sensitivity under low-data setting. Our code is available at \url{https://github.com/Muhammad-Al-Barbary/CoMoTo}.

\keywords{Breast Cancer  \and Knowledge Distillation \and Object Detection.}
\end{abstract}

\section{Introduction}

Breast cancer is the most prevalent type of cancer among women~\cite{intro_ref1}. Early detection of breast cancer significantly improves survival rates~\cite{intro_ref2}. Mammography, the standard screening protocol, is limited by its lack of depth perception which could lead to lesion masking by the surrounding tissue. Digital Breast Tomosynthesis (DBT), an innovative breast imaging modality, addresses this limitation by providing a pseudo 3D image that enables visualization of superimposed structures~\cite{intro_ref3}. Despite the higher detection rates, the 3D nature of DBT requires a much longer reading time, limiting its applicability in breast screening~\cite{intro_ref4,intro_ref5}.
Deep learning has been proposed by numerous studies for breast cancer detection in mammograms to enhance the efficiency of radiologists~\cite{intro_ref7,intro_ref8,intro_ref9,intro_ref10}. Nevertheless, applying these approaches to DBT is more challenging due to the scarcity of DBT data, which can be largely attributed to the extensive time required for reading and annotating images~\cite{intro_ref11}. Therefore, developing automated methods for detecting DBT lesions becomes crucial.

Recent studies have already addressed these issues by exploiting mammography data to enhance DBT breast cancer detection. For instance, various works have demonstrated that pretraining on mammography data, followed by fine-tuning on DBT, substantially enhances DBT lesion detection~\cite{intro_ref12,intro_ref9,intro_ref13}. While these methods outperform random initialization or natural images pretraining, they are notably vulnerable to catastrophic forgetting~\cite{intro_ref14}. 

Knowledge distillation (KD) emerges as a solution aimed at transferring knowledge from a teacher network to a student network by aligning their outputs at selected layers during training, minimizing the risk of catastrophic forgetting~\cite{intro_ref15}. KD has shown great potential in cross-modal medical applications, including retinal disease classification from Fundus to OCT imaging~\cite{intro_ref16}, lymph node metastasis prediction~\cite{intro_ref17}, and unpaired CT-MRI segmentation~\cite{dou2020ummkd}. Nevertheless, to the best of our knowledge, the application of KD to enhance lesion detection from DBT using mammography data remains unexplored.

Most existing cross-modal approaches heavily rely on image-level KD~\cite{intro_ref16,intro_ref17}. However, image-level KD can be hindered by the distilled information misalignment. Specifically, when applied to our task of unpaired mammography to DBT object detection, image-level KD performs poorly, especially in limited data setting (see~\cref{tab:main-results}). In object detection, image-level KD shows limited performance primarily due to its bias towards larger objects that dominate a substantial area of the image~\cite{intro_ref18}. This issue is further exacerbated when distilling knowledge from a 2D to a 3D modality, as image-level KD does not adequately address variances in local structures.

To this end, we propose Unpaired \textbf{C}r\textbf{o}ss-\textbf{Mo}dal Lesion Distillation Improves Breast Lesion Detection in \textbf{To}mosynthesis (CoMoTo). 
Specifically, we propose two novel components, \textbf{Lesion-specific KD (LsKD)} and \textbf{Intra-modal Point Alignment (ImPA)}. LsKD distills lesion-specific features from a mammography teacher to a DBT student, selectively focusing on relevant features while disregarding background information. In addition, LsKD efficiently manages the variability in lesion sizes by extracting a predetermined set of critical lesion points. ImPA further enriches LsKD by ensuring the alignment of critical features within the teacher before distilling knowledge to the student. Our extensive experimental evaluation at different classification levels and various data availability scenarios reflects the great potential of CoMoTo to improve breast lesion detection from DBT images under low data settings.

\section{Methodology}
\begin{figure}[t]
    \centering
    \includegraphics[width=1\linewidth]{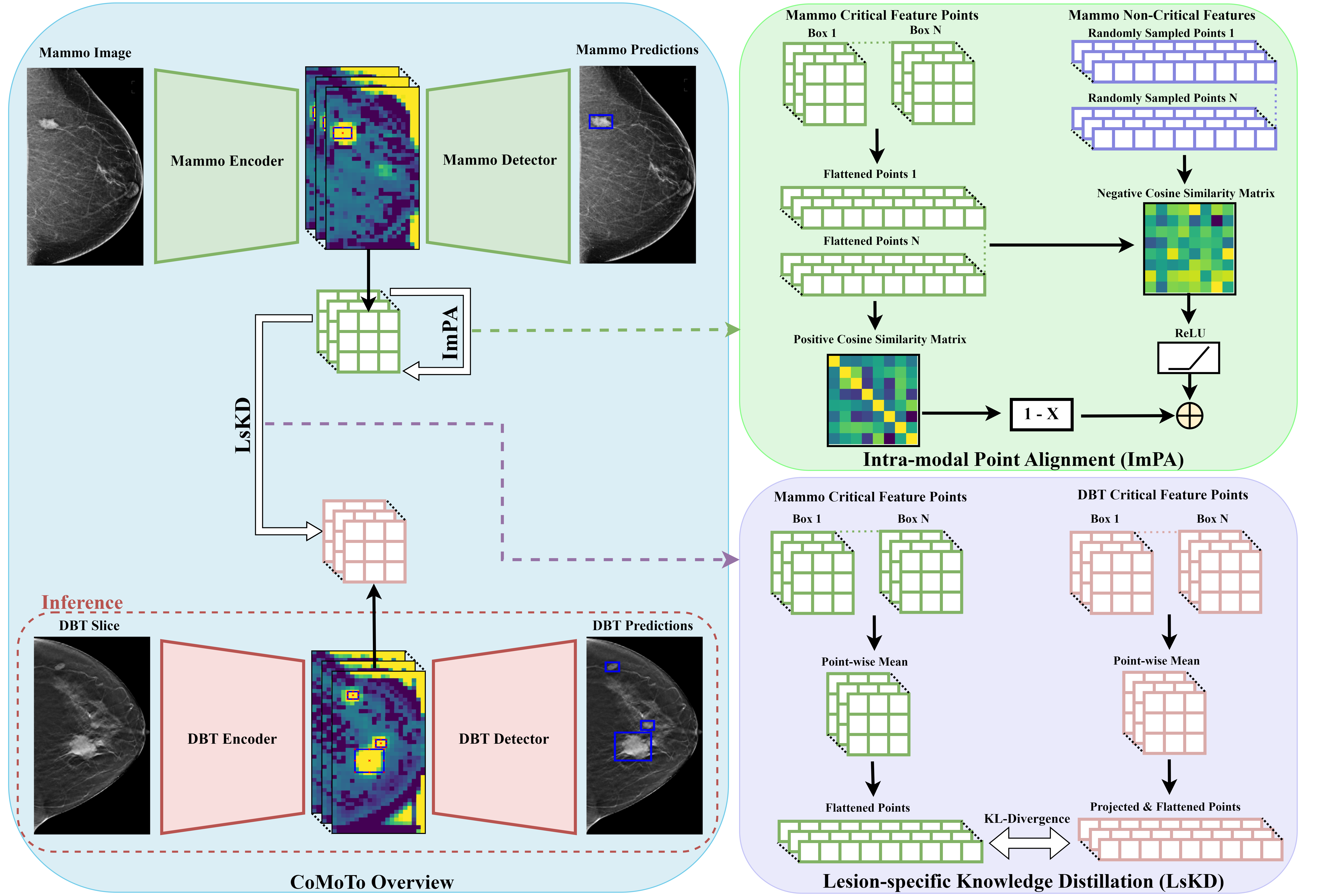}
    \caption{CoMoTo Overview. Feature maps are extracted from the encoders of modality dedicated object detection networks. Mammography critical lesion features corresponding to target bounding boxes are aligned with each other. Subsequently, DBT critical lesion features are aligned to those of mammography. During inference, the DBT model is used alone, improving efficiency and practicality.}
    \label{framework-overview}
\end{figure}

Our framework, CoMoTo, is based on the distillation of selected features corresponding to the object of interest from an intra-modal point-aligned mammography teacher network to a DBT student one. Fig.~\ref{framework-overview} shows an overview of the main components of CoMoTo. During training, a mammography dedicated network is trained over mammography data by minimizing a typical detection loss $\mathcal{L}_{det}$ and Intra-modal Point Alignment (ImPA) loss $\mathcal{L}_{ImPA}$, which maximizes the similarity of critical lesion features of different mammography samples. Similarly, a DBT dedicated network is trained over DBT data to minimize $\mathcal{L}_{det}$. Additionally, KD is applied by minimizing Lesion-specific KD (LsKD) loss $\mathcal{L}_{LsKD}$,
increasing the similarity of the DBT critical lesion features extracted from the DBT encoder with the mammography intra-modal point aligned critical lesion features extracted from the mammography encoder. During inference, only the DBT network is used to detect lesions from DBT data.

In the next sections, we introduce our proposed Lesion-specific KD by critical feature points alignment, then we introduce Intra-modal Point Alignment of the teacher to further enhance Lesion-specific KD.

\subsection{Cross-modal Lesion-specific Knowledge Distillation} \label{LsKD}

Existing KD frameworks aim to distill knowledge from a pretrained teacher model to a student model through image-level feature alignment~\cite{intro_ref16,intro_ref17,method_ref1}. 
However, within the realm of breast imaging, this approach proves to be highly susceptible to the global variations inherent in breast tissue. Moreover, image-level feature alignment methods exhibit a bias towards larger objects, which inherently occupy larger regions of the feature maps~\cite{intro_ref18}. To address these challenges, we propose Lesion-specific KD.

Aligning lesion-specific features is not straightforward, since lesions can vary in size and structure, which limits the applicability of traditional features distillation. 
Therefore, we propose to identify 9 critical points in the extracted feature maps of both modalities for each existing object, representing its center, 4 edges, and 4 sides midpoints. This is performed by scaling the ground truth bounding boxes to correspond to the down-sampled feature maps. 
Afterwards, for each bounding box, its scaled 9 critical points are employed to select 9 features from the extracted feature maps, resulting in flattened critical feature vectors ${f_{c}}\in \mathbb{R}^{D \times 9}$, where $D$ is the number of feature maps channels. 
Point-wise mean is then applied for each modality to average the critical feature vectors of the whole batch, creating mammography and DBT lesion-specific prototypes. The derivation of the lesion-specific prototype of one modality ${e}_{c}$ can be formulated as:
\begin{equation} 
\boldsymbol{e}_{c}=\frac{1}{B} \cdot \sum_{i=1}^{B} \frac{1}{k_i} \cdot \sum_{j=1}^{k_i}f_{\text{c,i,j}}
\end{equation} 
where $B$ is the number of images per batch, and $k_i$ is the number of ground truth lesions in one image. Softmax activation $\sigma$ is then applied to both prototypes:  
 \begin{equation}
\mathcal{E}_{c,mammo}=\sigma(\frac{\boldsymbol{e}_{c,mammo}}{\tau}), \mathcal{E}_{c,dbt}=\sigma(\frac{\mathbf{P}(\boldsymbol{e}_{c,dbt})}{\tau}) 
\label{eq:softmax}
\end{equation}
where $\tau$ is a temperature parameter to soften the prototypes distribution, and $\mathbf{P}$ is a learnable linear layer that projects the DBT prototype, reducing the domain shift of the DBT and mammography. Finally, the two lesion-specific prototypes are aligned by minimizing KL-divergence loss function:

\begin{equation} 
\mathcal{L}_{LsKD}= \frac{1}{D}\sum_{i=1}^D{\sum_{j=1}^9{{\mathcal{E}_{c,mammo,i,j}\log{\left(\frac{\mathcal{E}_{c,mammo,i,j}}{\mathcal{E}_{c,dbt,i,j}}\right)}}}}
\end{equation}
The overall loss function to be optimized can be then defined as:
\begin{equation} 
\mathcal{L}_{DBT} = \mathcal{L}_{det} + \alpha \cdot\mathcal{L}_{LsKD}
\label{eq:distillation}
\end{equation} 
where $\alpha$ is a loss weight to control the contribution of lesion-specific KD.

\subsection{Intra-modal Point Alignment}

Our proposed lesion-specific KD assumes homogeneous teacher features across different lesions. Thus, the presence of non-homogeneous point-wise teacher features has the potential to adversely impact performance. To this end, we propose Intra-modal Point Alignment to explicitly align critical lesions feature points between different mammography images during training of the mammography network. 

During the mammography teacher training, critical feature vectors ${f_{c}}$ of each lesion in the batch are extracted, as described in Sec. \ref{LsKD}. Additionally, non-critical feature vectors of the same dimensions ${f_{nc}}\in \mathbb{R}^{D \times 9}$ are extracted by randomly sampling the background features, which are not part of the scaled bounding boxes, from each image in the batch. We design $\mathcal{L}_{ImPA}$ to include a positive term that maximizes the cosine similarity between critical features from different data samples, and a negative term that penalizes positive cosine similarities between critical and non-critical features, preventing model collapse. Our proposed loss can be formulated as: 
\begin{equation}
\mathcal{L}_{ImPA} = \mathcal{L}_{pos} + \mathcal{L}_{neg}
\end{equation}
$\mathcal{L}_{pos}$ and $\mathcal{L}_{neg}$ can accordingly be formulated as: 
\begin{equation}
\mathcal{L}_{pos} = \frac{1}{B^2}\sum_{i=1}^{B}\sum_{j=1}^{B} 1 - \frac{{f_{c, i}} \cdot {f_{c, j}}}{\|{f_{c,i}}\| \|{f_{c, j}}\|}
\end{equation}
\begin{equation}
\mathcal{L}_{neg} = \frac{1}{B^2}\sum_{i=1}^{B}\sum_{j=1}^{B} max(0, \frac{{f_{c, i}} \cdot {f_{nc, j}}}{\|{f_{c,i}}\| \|{f_{nc, j}}\|})
\end{equation}
The overall mammography loss, $\mathcal{L}_{mammo}$, is the result of combining the detection loss with the Intra-modal Point Alignment loss as follows:
\begin{equation}
\mathcal{L}_{mammo} = \mathcal{L}_{det} + \beta \cdot \mathcal{L}_{ImPA} 
\label{eq:alignment}
\end{equation}
where $\beta$ is a loss weight that controls the contribution of Intra-modal Point Alignment during mammography training.

\section{Experiments}
\subsection{Experimental Setup}
\subsubsection{Datasets.} 
To build and evaluate our framework, we used two unpaired datasets for mammography and DBT. The OPTIMAM Medical Image Database (\textbf{OMI-DB}) is a huge, multi-centre, dataset consisting of mammography images and associated clinical information from UK~\cite{dataset_ref1}. For the sake of this study, we select a subset of images with available bounding box coordinates representing identified masses.  The subset dataset consists of 3608 full-field digital mammography images from 1911 patients. This dataset comprises images of both right and left breasts, captured from craniocaudal and mediolateral oblique views. It's worth noting that some patients may have one or more of the four image views missing. The image resolution ranged from $1809\times551$ to $4084\times3328$. This dataset included 3608 lesions.

For DBT, we used a subset of the Duke University Breast Cancer Screening DBT (\textbf{BCS-DBT})~\cite{dataset_ref2} that was used in the DBTex challenge~\cite{dataset_ref3}. Similar to mammography, we selected the volumes with available annotations representing masses. Overall, our used dataset consisted of 410 volumes from 396 patients including volumes of right and left breasts, from craniocaudal and mediolateral oblique views, with some patients having one or more of the four volumes missing. The volume resolution ranged from $1465\times 454\times 24$ to $2457\times 1916\times 114$. Overall, the used dataset included 431 lesions.

\noindent{\textbf{Implementation Details.}} For mammography preprocessing, the dataset was stratified by patient and divided into training, validation, and testing sets of approximately 60\%, 20\%, and 20\%, respectively. We cropped the breast regions and resized them to $2048\times1024$. During training, we applied random horizontal flipping, ensuring homogeneous performance across both breasts. 
For DBT, following the DBTex challenge~\cite{dataset_ref3} distribution, the dataset was divided into 50\%, 20\%, and 30\% for training, validation, and testing, respectively. We sliced the volumes into 2D images and used the central images of lesions to train the DBT model. The same mammography preprocessing and augmentation were applied to the DBT. During evaluation, non-max suppression was applied to the DBT slices predictions of each volume.

Our framework is applicable to any architecture that include a feature extractor. In our experiments, we used RetinaNet as the detection network for both modalities with Focal Loss~\cite{implement_ref1} as $\mathcal{L}_{det}$ (see \cref{eq:distillation,eq:alignment}). We used ResNet50~\cite{implement_ref2} as the feature extractor, anchor sizes of 32, 64, 128, 256, and 512, and aspect ratios of 0.5, 1, and 2. We used Stochastic Gradient Descent as an optimization algorithm, with a learning rate of 0.01 and a momentum of 0.9. A batch size of 8 images was used. We selected $\tau = 4$ in \cref{eq:softmax}, $\alpha = 2$ in \cref{eq:distillation}, and $\beta = 2$ in \cref{eq:alignment}. The models were trained for 50 epochs in all experiments, and the learning rate was divided by 10 after 30 epochs. The validation mean Average Precision (mAP) was the used criterion to select the best checkpoint. The framework was implemented on a Nvidia A40 and a Nvidia A30 GPUs.

\noindent{\textbf{Evaluation Metrics.}} Following the DBTex challenge~\cite{dataset_ref3}, our primary metric was mean sensitivity at 1, 2, 3, and 4 false positives per case (Mean Sensitivity), and our secondary metric was sensitivity at 2 false positives (Sensitivity@2FPs).

\subsection{Results}

\begin{figure}[t]
    \centering
    \includegraphics[width=1\linewidth]{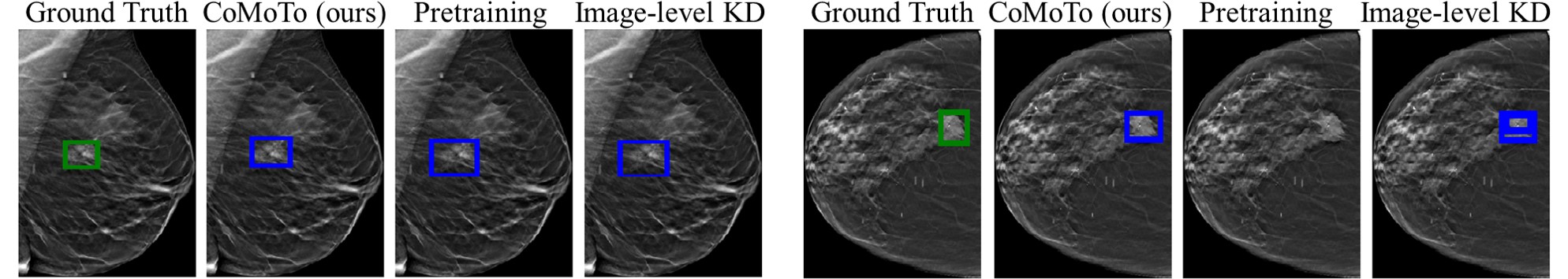}
    \caption{Qualitative assessment on testing data shows CoMoTo's more accurate lesion fitting compared to other SOTA approaches when trained over 10\% of the data.}
    \label{qualitative_assessment}
\end{figure}

Our experiments, as shown in Tab.~\ref{tab:main-results}, show that mammography pretraining followed by DBT fine-tuning, as in~\cite{intro_ref12}, could achieve 0.57 and 0.64 Mean Sensitivity when trained over 10\% and 25\% of the available DBT data. However, applying the SOTA image-level KD, as in~\cite{intro_ref16}, resulted in a performance downgrade by 3\% and 5\% Mean Sensitivity, when using 10\% and 25\% of the data, respectively. Applying image-level KD from an unpaired 2D modality, like mammography, to a 3D modality, such as DBT, introduces undesirable noise as a result of distilling misaligned information. CoMoTo overcomes these limitations by distilling lesion-specific features and ignoring misaligned background features, resulting in a Mean Sensitivity improvement by 7\% and 10\%, when using 10\% of the data, compared to pretraining and image-level KD, respectively. Additionally, the improvement is also observable when training the model over 25\% of the data, where CoMoTo achieves 3\% and 9\% higher Mean Sensitivity compared to pretraining and image-level KD, respectively. The qualitative assessment, illustrated in Fig.~\ref{qualitative_assessment}, also highlights CoMoTo's ability to decrease error rates and enhance the detection window fitting, compared to other SOTA approaches.

\begin{table}[t]
\caption{CoMoTo outperforms pretraining and image-level KD under low data setting.}
\label{tab:main-results}
\centering
\begin{adjustbox}{width=1\textwidth}
\begin{tabular}{c|cc|cc}
\hline
\multirow{2}{*}{Method}   & \multicolumn{2}{c|}{10\% training data}  & \multicolumn{2}{c}{25\% training data}\\
\cline{2-5}
 & \textbf{Mean Sensitivity}  &  \textbf{Sensitivity@2FPs}  & \textbf{Mean Sensitivity}  &  \textbf{Sensitivity@2FPs}\\
\hline
Pretraining~\cite{intro_ref12} & 0.570 & 0.551 & 0.640 & 0.618 \\
Image-level KD~\cite{intro_ref16} & 0.543 & 0.536 & 0.586 & 0.563 \\
\textbf{CoMoTo (ours)} & \textbf{0.639}  & \textbf{0.603} & \textbf{0.674} & \textbf{0.640}\\
\hline
\end{tabular}
\end{adjustbox}
\end{table}

\begin{table}
\caption{Effects of LsKD and ImPA. Disabling LsKD is associated with applying image-level distillation \cite{intro_ref16} from a normal mammography pretrained model, as in first row, or an ImPA pretrained model, as in   third row.
}
\label{tab:intra-modal}
\centering
\resizebox{0.7\textwidth}{!}{%
\begin{tabular}
{cc|c|cc}
\hline
\multicolumn{2}{c|}{\textbf{Method}} & \multirow{2}{*}{\textbf{Distillation}} & \multirow{2}{*}{\textbf{Mean Sensitivity}} & \multirow{2}{*}{\textbf{Sensitivity@2FPs}} \\
\cline{0-1}
\textbf{ImPA} & \textbf{LsKD} & & &   \\
\hline
   \multicolumn{5}{c}{10\% training data} 
  \\
  \hline
\ding{55} & \ding{55} & image-level & 0.543 & 0.536 \\
\ding{55} & \ding{51} & lesion-level & 0.563 & 0.563 \\
\ding{51} & \ding{55} & image-level & \underline{0.629} & \textbf{0.610} \\
\ding{51} & \ding{51} & lesion-level & \textbf{0.639} &  \underline{0.603} \\
\hline
   \multicolumn{5}{c}{25\% training data} 
  \\
  \hline
\ding{55} & \ding{55} & image-level & 0.586 & 0.563 \\
\ding{55} & \ding{51} & lesion-level & 0.638 & 0.625 \\
\ding{51} & \ding{55} & image-level & \underline{0.661} & \underline{0.632} \\
\ding{51} & \ding{51} & lesion-level & \textbf{0.674} &  \textbf{0.640} \\
\hline
\end{tabular}
}
\vspace{-0.65cm}
\end{table}

\subsection{Ablation Studies}

\begin{figure}[ht]
    \centering
    \includegraphics[width=1\linewidth]{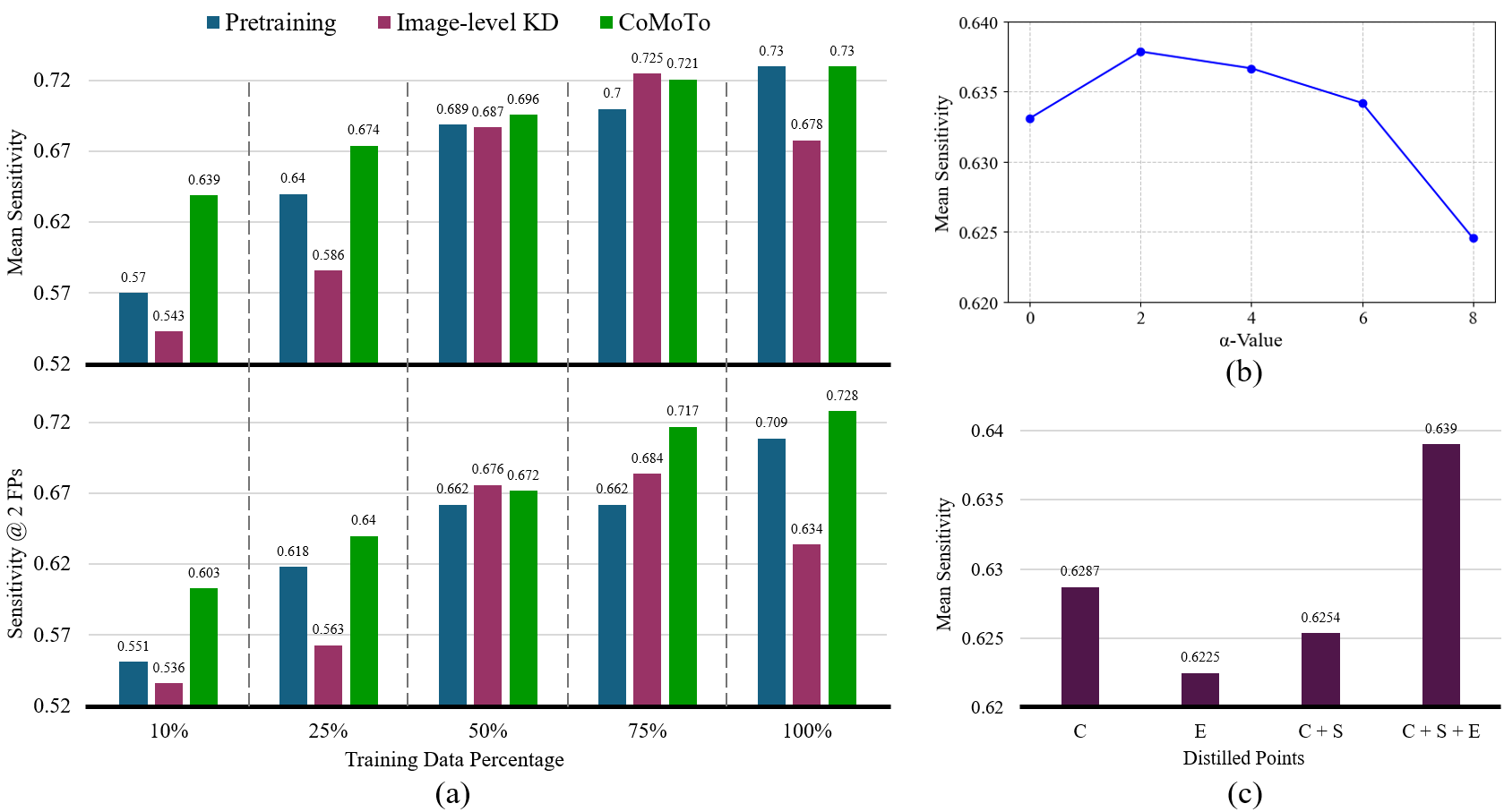}
    \caption{(a) Comparing CoMoTo with SOTA at varying data ratios. (b) $\alpha$ effect on lesion-specific KD. (c) Effect of different distillation points. C, E, and S refer to the center, edges, and sides midpoints of the lesion bounding boxes, respectively.}
    \label{ablations}
\end{figure}

\subsubsection{CoMoTo Component Analysis.} We assess the effect of each of our modules, as shown in Tab.~\ref{tab:intra-modal}. LsKD outperforms the SOTA image-level KD~\cite{intro_ref16}, in terms of Mean Sensitivity, by $2\%$ and $5\%$, whereas ImPA has a larger impact improving by $9\%$ and $8\%$ over the image-level KD and by $7\%$ and $2\%$ using LsKD without previous ImPA, when using 10\% and 25\% DBT training data, respectively. Those results show the impact of aligning the lesion features of both the student and teacher networks, an aspect often overlooked in the literature.

\noindent{\textbf{Training Data Ratio Effect.}} We compare mammography pretraining~\cite{intro_ref12} and image-level KD~\cite{intro_ref16} with our framework, CoMoTo, at different training data ratios as shown in Fig.~\ref{ablations} (a). The results indicate that CoMoTo is superior to other SOTA methods, especially at lower data setting.

\noindent{{\(\boldsymbol{\alpha}\)\textbf{-value Effect.}}} Fig.~\ref{ablations} (b) shows the results of our method at different values of $\alpha$ when trained over 10\% of the DBT data. The results reveal that significantly raising alpha can cause the model to deviate from the detection objective.

\noindent{{\textbf{Distilled Points.}}} As depicted in Fig.~\ref{ablations} (c), distilling 9 critical points, representing the center, sides midpoints, and edges, improves the overall KD process, compared to distilling other sets of points. This suggests that combining features associated with varying degrees of certainty enhances the value of distilled knowledge.

\section{Conclusion}

Our work proposes a novel KD framework, CoMoTo, to improve breast lesion detection from DBT. CoMoTo aligns lesions features extracted from a mammography network across different mammography samples, then distills the aligned lesions features from mammography to DBT, while ignoring background features corresponding to unpaired breast tissues. Moreover, CoMoTo remains unbiased towards larger lesions by distilling a predetermined number of critical points from each lesion, addressing a significant limitation of image-level KD.
Our experiments demonstrate that CoMoTo outperforms existing approaches by a considerable margin, especially at low data setting. The results of this study reveal a great potential of object-level KD to improve object detection in a less available modality using unpaired data from a more common one.

\vspace{6pt}
\noindent
\textbf{Acknowledgement.} This work was partially funded by the Erasmus+: Erasmus Mundus Joint Master’s Degree (EMJMD) scholarship with project reference 610600-EPP-1-2019-1-ES-EPPKA1-JMD-MOB and the project VICTORIA, “PID2021-123390OB-C21” from Ministerio de Ciencia e Innovación of Spain.

\bibliographystyle{splncs04}
\bibliography{ref}

\end{document}